# A Deep Learning Framework for Assessing Physical Rehabilitation Exercises

Yalin Liao, Aleksandar Vakanski, *Member, IEEE,* and Min Xian, *Member, IEEE* [1]

*Abstract*—Computer-aided assessment of physical rehabilitation entails evaluation of patient performance in completing prescribed rehabilitation exercises, based on processing movement data captured with a sensory system. Despite the essential role of rehabilitation assessment toward improved patient outcomes and reduced healthcare costs, existing approaches lack versatility, robustness, and practical relevance. In this paper, we propose a deep learning-based framework for automated assessment of the quality of physical rehabilitation exercises. The main components of the framework are metrics for quantifying movement performance, scoring functions for mapping the performance metrics into numerical scores of movement quality, and deep neural network models for generating quality scores of input movements via supervised learning. The proposed performance metric is defined based on the log-likelihood of a Gaussian mixture model, and encodes low-dimensional data representation obtained with a deep autoencoder network. The proposed deep spatio-temporal neural network arranges data into temporal pyramids, and exploits the spatial characteristics of human movements by using sub-networks to process joint displacements of individual body parts. The presented framework is validated using a dataset of ten rehabilitation exercises. The significance of this work is that it is the first that implements deep neural networks for assessment of rehabilitation performance.

## I. Introduction

PARTICIPATION in physical therapy and rehabilitation programs is often compulsory and critical in postoperative recovery or for treatment of a wide array of musculoskeletal conditions. However, it is infeasible and economically unjustified to offer patient access to a clinician for every single rehabilitation session [1]. Accordingly, current healthcare systems around the world are organized such that an initial portion of rehabilitation programs is performed in an inpatient facility under direct supervision by a clinician, followed by a second portion performed in an outpatient setting, where patients perform a set of prescribed exercises in their own residence. Reports in the literature indicate that more than 90% of all rehabilitation sessions are performed in a home-based setting [2]. Under these circumstances, patients are tasked to record their daily progress and periodically visit the clinic for functional assessment. Still, numerous medical sources report low levels of patient adherence to the recommended exercise regimens in home-based rehabilitation, leading to prolonged treatment times and increased healthcare cost [3], [4]. Although many different factors have been identified that contribute to the low compliance rates, the major impact factor is the absence of continuous feedback and oversight of patient exercises by a healthcare professional [5]. Despite the development of a variety of tools and devices in support of physical rehabilitation, such as robotic assistive systems [6], virtual reality and gaming interfaces [7], and Kinect-based assistants [2], there is still a lack of versatile and robust systems for automatic monitoring and assessment of patient performance.

The article proposes a novel framework for assessment of home-based rehabilitation that encompasses formulation of metrics for quantifying movement performance, scoring functions for mapping the performance metrics into numerical scores of movement quality, and deep learning-based end-to-end models for encoding the relationship between movement data and quality scores. The employed performance metric is based on probabilistic modeling of the skeletal joints data with a Gaussian mixture model, and consequently, it employs the log-likelihood of the model for performance evaluation [8]. Next, the article investigates the effectiveness of deep autoencoder neural networks for dimensionality reduction of captured data. Further, we propose a scoring function for scaling the values of the performance metric into movement quality scores in the [0, 1] range. The resulting scores are employed as the ground truth for training the proposed deep neural networks (NNs) for rehabilitation applications.

The paper introduces a deep NN model designed to handle spatial and temporal variability in human movements. Motivation for the proposed network structure was prior work on temporal pyramids [9] and hierarchical recurrent networks for motion classification [10]. Specifically, the proposed model aims to exploit spatial characteristics of human movements by hierarchical processing of the joint displacements of different body parts via a series of sub-networks that gradually merge the extracted feature vectors. Temporal pyramids are introduced using movement sequences at different time scales in order to learn data representations at multiple levels of abstraction. The network contains both convolutional layers for learning spatial dependencies and recurrent layers for encoding temporal correlations in movement data. The framework is validated on

[1] Manuscript submitted June 14, 2019. This work was supported by the Center for Modeling Complex Interactions (CMCI) at the University of Idaho through NIH Award #P20GM104420.

Yalin Liao, Aleksandar Vakanski, and Min Xian are with the Department of Computer Science, University of Idaho, 1776 Science Center Drive, Idaho Falls, ID, 83402, USA (e-mail: liao4728@vandals.uidaho.edu; vakanski@uidaho.edu; mxian@uidaho.edu).

2the University of Idaho – Physical Rehabilitation Movement Dataset (UI-PRMD) [11]. To the best of our knowledge, this is the first framework that employs deep NNs for assessment of rehabilitation exercises.

The main contributions of the paper are: (1) A novel framework for computer-aided assessment of rehabilitation exercises; (2) A deep spatio-temporal NN model for outputting movement quality scores; and (3) A performance metric that employs probabilistic modeling and autoencoder NNs for dimensionality reduction of rehabilitation data.

The article is organized as follows. The next section provides an overview of related work. Section III first introduces the mathematical notation and afterward describes the components of the proposed framework for rehabilitation assessment, including dimensionality reduction, performance metric, scoring function, and deep learning model. The validation of the proposed framework on a dataset of rehabilitation exercises is presented in Section IV. The last two sections discuss the results and summarize the paper.

## II. RELATED WORK

### A. Human Movement Modeling

Conventional approaches for mathematical modeling and representation of human movements are broadly classified into two categories: top-down approaches that introduce latent states for describing the temporal dynamics of the movements, and bottom-up approaches that employ local features for representing the movements. Commonly used methods in the first category include Kalman filters [12], hidden Markov models [13], and Gaussian mixture models [14]. The main shortcomings of these methods originate from employing linear models for the transitions among the latent states (as in Kalman filters), or from adopting simple internal structures of the latent states (typical for hidden Markov models). The approaches based on extracting local features employ predefined criteria for identifying key points [15] and/or required body postures [16], [17], or a collection of statistics of the movements (e.g., mean, standard deviation, mode, median) [18]. Such local features are typically motion-specific, which limits the ability to efficiently handle arbitrary spatio-temporal variations within movement data.

Recent developments in artificial NNs stirred significant interest in their application for modeling and analysis of human motions. Numerous works employed NNs for motion classification and applied the trained models for activity recognition, gait identification, gesture recognition, action localization, and related applications. NN-based motion classifiers utilizing different computational units have been proposed, including convolutional units [19], long short-term memory (LSTM) recurrent units [20], gated recurrent units, and combinations [21] or modifications of these computational units [22]. Also, NNs with different layer structures have been implemented, such as encoder-decoder networks, spatio-temporal graphs [23], and attention mechanism models [24]. Besides the task of classification, a body of work in the literature focused on modeling and representation of human movements for prediction of future motion patterns [25], synthesis of movement sequences [26], and density estimation [8]. Conversely, little research has been conducted on the application of NNs for evaluation of movement quality, which can otherwise find use in various applications (physical rehabilitation being one of them).

### B. Movement Assessment

Quantifying the level of correctness in completing prescribed exercises is important for the development of tools and devices in support of home-based rehabilitation. The movement assessment in existing studies is typically accomplished by comparing a patient's performance of an exercise to the desired performance by healthy participants.

Several studies in the literature on exercise evaluation employed machine learning methods to classify the individual repetitions into correct or incorrect classes of movements. Methods used for this purpose include Adaboost classifier, $k$-nearest neighbors, Bayesian classifier, and an ensemble of multi-layer perceptron NNs [27]–[29]. The outputs in these approaches are discrete class values of 0 or 1 (i.e., incorrect or correct repetition). However, these methods do not provide the capacity to detect varying levels of movement quality or identify incremental changes in patient performance over the duration of the rehabilitation program.

The majority of related studies employ distance functions for deriving movement quality scores. Concretely, Houmanfar et al. [18] used a variant of the Mahalanobis distance to quantify the level of correctness of rehabilitation movements, based on a calculated distance between patient-performed repetitions and a set of repetitions performed by a group of healthy individuals. Similarly, a body of work utilized the dynamic time warping (DTW) algorithm [30] for calculating the distance between a patient's performance and healthy subjects' performance [31]–[33]. The advantage of the distance functions is that they are not exercise-specific, and thus can be applied for assessment of new types of exercises. The distance functions also have shortcomings, because they do not attempt to derive a model of the rehabilitation data, and the distances are calculated at the level of individual time-steps in the raw sensory measurements.

A line of research utilized probabilistic approaches for modeling and evaluation of rehabilitation movements. Studies based on hidden Markov models [34], [35] and mixtures of Gaussian distributions [8] typically perform a quality assessment based on the likelihood that the individual sequences are being drawn from a trained model. Although the utilization of probabilistic models is advantageous in handling the variability due to the stochastic character of human movements, models with abilities for a hierarchical data representation can produce more reliable outcomes for movement quality assessment, and better generalize to new exercises.

## III. PROPOSED METHOD

A block-diagram of the envisioned framework for assessing rehabilitation exercises is depicted in Fig. 1. The skeletal joint coordinates acquired by the sensory system are processed via dimensionality reduction, performance quantification, and scoring mapping to obtain movement quality scores that are subsequently used for training an NN model. The trained NN model is afterward used to automatically generate movement quality scores for input movement data acquired by the sensory system.

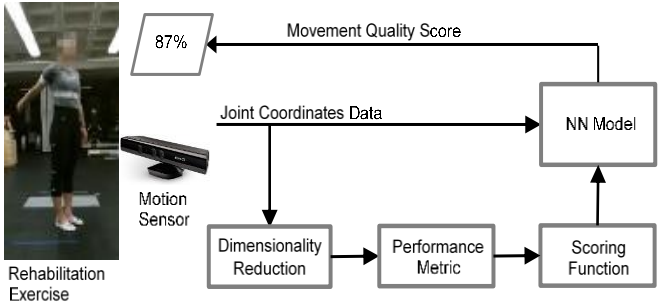

Fig. 1. Overview of the proposed framework for assessment of rehabilitation exercises.

*A. Notation*

In outpatient physical rehabilitation, a daily rehabilitation session requires completing a series of exercises, where the patient is instructed to complete a certain number of repetitions of each exercise during each session. The acquired data by the sensory system for one particular exercise performed by $S$ healthy subjects is denoted by $\mathbb{X}$, and hereafter they are referred to as reference movements. The symbol $R_s$ is used for the number or repetitions of the exercise by the $s$-th subject. The combined data for all $R_s$ repetitions of the exercise by the $s$-th subject is denoted $\Delta_s$. Similar, $R$ is used for the total number of all repetitions by the $S$ subjects, i.e., $R = \sum_{s=1}^{S} R_s$. Using the notation $\mathbf{X}_{s,r}$ for the collected data of the $r$-th repetition by the $s$-th subject, we have $\mathbb{X} = \{\Delta_s\}$ for $s \in \underline{S}$, where $\Delta_s = \{\mathbf{X}_{s,r}\}$ for $r \in \underline{R_s}$. For convenience, throughout the text the underscore symbol denotes a set of indices, e.g., $\underline{S} = \{1, 2, \dots, S\}$ for any positive integer $S$. The data for each repetition $\mathbf{X}_{s,r}$ is a temporal sequence of $T$ measurements, therefore $\mathbf{X}_{s,r} = (\mathbf{x}_{s,r}^{(1)}, \mathbf{x}_{s,r}^{(2)}, \dots, \mathbf{x}_{s,r}^{(T)})$, where the superscripts are used for indexing the temporal order of the joint displacement vectors within the repetition. Furthermore, the individual measurement $\mathbf{x}_{s,r}^{(t)}$ for $t \in \underline{T}$ is a $D$-dimensional vector, consisting of the values for all joint displacements in the human body, i.e. $\mathbf{x}_{s,r}^{(t)} = [x_{s,r}^{(t,1)}, x_{s,r}^{(t,2)}, \dots x_{s,r}^{(t,D)}]$.

The collected data for the patient group are referred in the article as patient movements, and are denoted with the symbol $\mathbb{Y}$. By analogy to the introduced notation for the reference movements, $\mathbb{Y} = \{\mathbf{Y}_{s,r}\}$, where $\mathbf{Y}_{s,r}$ is the data of the $r$-th repetition by the $s$-th subject. Analogously, the repetition $\mathbf{Y}_{r,s} = (\mathbf{y}_{s,r}^{(1)}, \mathbf{y}_{s,r}^{(2)}, \dots, \mathbf{y}_{s,r}^{(T)})$ is comprised of a sequence of multidimensional vectors $\mathbf{y}_{s,r}^{(t)} = [y_{s,r}^{(t,1)}, y_{s,r}^{(t,2)}, \dots, y_{s,r}^{(t,D)}]$.

*B. Dimensionality Reduction*

The sensory systems for motion capturing typically track between 15 and 40 skeletal joints, depending on the sensor type. The measurement data consists of 3-dimensional spatial positions and/or orientations for each joint, and therefore the dimensionality of the data ranges between 45 and 120. Dimensionality reduction of recorded data is an essential step in processing human movements to suppress unimportant, redundant, or highly correlated dimensions. The aim is to project the data $\mathbb{X} = \{\mathbf{X}_{s,r} = (\mathbf{x}_{s,r}^{(t)}): \mathbf{x}_{s,r}^{(t)} \in \mathbb{R}^D\}$ into a lower-dimensional representation $\widetilde{\mathbb{X}} = \{\widetilde{\mathbf{X}}_{s,r} = (\tilde{\mathbf{x}}_{s,r}^{(t)}): \tilde{\mathbf{x}}_{s,r}^{(t)} \in \mathbb{R}^M\}$, for $t \in \underline{T}$, $s \in \underline{S}$, $r \in \underline{R_s}$, where $M < D$.

A common approach for dimensionality reduction of human movement data is maximum variance [36], which simply retains the first $M$ dimensions with the largest variance and discards the remaining dimensions. Principal component analysis (PCA) and its variants [37] are also widely used for reducing the dimensionality of movement data, where a matrix containing the leading $M$ eigenvectors corresponding to the largest eigenvalues of the covariance matrix $\mathbf{V}$ is used for projecting the data into a lower-dimensional space. Although PCA is one of the most common approaches for dimensionality reduction in general, it employs linear mapping of high-dimensional data into a lower-dimensional representation. Likewise, the shortcomings of maximum variance originate from its simplicity.

In the proposed framework, we introduce autoencoder NNs [38] for dimensionality reduction. Autoencoder NN is a nonlinear technique for dimensionality reduction allowing extracting richer data representations for dimensionality reduction in comparison to the linear techniques (such as PCA). Furthermore, deep autoencoder NNs created by stacking multiple consecutive layers of hidden neurons, can additionally increase the representational capacity of the network.

Autoencoders are used for unsupervised learning of an alternative representation of input data, through a process of data compression and reconstruction. The data processing involves an encoding step of compressing input data through one or multiple hidden layers, followed by a decoding step of reconstructing the output from the encoded representation through one or multiple hidden layers. If $\mathcal{A}$ denotes a class of mapping functions from $\mathbb{R}^M$ to $\mathbb{R}^D$, and $\mathcal{B}$ is a class of mapping functions from $\mathbb{R}^D$ to $\mathbb{R}^M$, then for any function $A \in \mathcal{A}$ and $B \in \mathcal{B}$, the encoder portion projects an input $\mathbf{x}_{s,r}^{(t)} \in \mathbb{R}^D$ into a lower-dimensional representation $\tilde{\mathbf{x}}_{s,r}^{(t)} = B(\mathbf{x}_{s,r}^{(t)}) \in \mathbb{R}^M$ (referred to as a code), and the decoder portion converts the code into an output $A\left(B(\mathbf{x}_{s,r}^{(t)})\right) \in \mathbb{R}^D$. Autoencoders are trained to find functions $A \in \mathcal{A}$ and $B \in \mathcal{B}$ which minimize the mean squared deviation between the input data and output data, i.e.,

$$\operatorname*{argmin}_{A,B} \left\| A\left(B(\mathbf{x}_{s,r}^{(t)})\right) - \mathbf{x}_{s,r}^{(t)} \right\|. \tag{1}$$

A graphical representation of the adopted architecture for the autoencoder network is presented in Fig. 2. The encoder portion consists of three intermediate layers of LSTM recurrent units with 30, 10, and 4 computational units, and the corresponding decoder portion has three intermediate layers of LSTM units with 10, 30, and 117 computational units, respectively. The input time-series data are 117-dimensional vectors of joint coordinates. The code representation of the proposed network is a temporal sequence of 4-dimensional vectors.

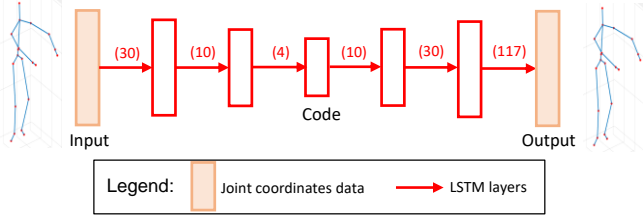

Fig. 2. The proposed autoencoder architecture projects an input movement data into a code representation, and re-projects the code into the movement data.

*C. Performance Metric*

The metrics for quantifying the patient performance are classified into model-less and model-based groups of metrics [39]. The model-less metrics employ distance functions, such as Euclidean, Mahalanobis distance, and dynamic time warping (DTW) [30] deviation between data sequences. The model-based metrics apply probabilistic approaches for modeling the movement data, and employ the log-likelihood for performance evaluation [8].

We adopt a metric based on Gaussian mixture model (GMM) log-likelihood. The choice stems from the demonstrated capacity of statistical methods to encode the inherent random variability in human movements; this results in improved ability by the model-based metrics to handle spatio-temporal variations in rehabilitation data. Log-likelihood of a movement data for a given model is a natural choice for evaluation of data instances in probabilistic models.

GMM is a parametric probabilistic model for representing data with a mixture of Gaussian probability density functions *[40]*. GMM is frequently used for modeling human movements. For a dataset consisting of multidimensional vectors $\boldsymbol{x}_{s,r}^{(t)}$, a GMM with $C$ Gaussian components has the form

$$\mathcal{P}(\boldsymbol{x}_{s,r}^{(t)}|\lambda) = \sum_{c=1}^{C} \pi_c \mathcal{N}(\boldsymbol{x}_{s,r}^{(t)}|\mu_c, \Sigma_c), \quad (2)$$

where $\lambda = \{\pi_c, \mu_c, \Sigma_c\}$ are the mixing coefficient, mean, and covariance of the *c*-th Gaussian component, respectively. The most popular method for estimating the model parameters $\lambda$ in GMM is the expectation maximization (EM) algorithm [41]; other approaches include maximum-a-posteriori estimation [42] and mixture density networks [40]. Subsequently, for a GMM model with parameters λ, the negative log-likelihood is used as a performance metrics, and for the repetition $\mathbf{Y}_{s,r}$ is calculated as

$$\mathcal{P}(\mathbf{Y}_{s,r}|\lambda) = -\sum_{t=1}^{T} \log\{\sum_{c=1}^{C} \pi_c \mathcal{N}(\mathbf{y}_{s,r}^{(t)}|\mu_C, \Sigma_C)\}. \quad (3)$$

*D. Scoring Function*

In the presented framework, a scoring function maps the values of the performance metrics into a movement quality score in the range between 0 and 1.

The resulting movement quality scores play a dual role in the framework. First, in a real-world exercise assessment setting, the quality scores allow for intuitive understanding of the calculated values of the used performance metric. For instance, a movement quality score of 88% presented to a patient is easy to understand, and it can also enable the patient to self-monitor his/her progress toward functional recovery based on received scores over a period of time. Second, the movement quality scores are used here for supervised training of the deep NN models.

For a sequence of performance metric values of the reference movements $\mathbf{x} = (x_1, x_2, \cdots, x_L)$ and a sequence $\mathbf{y} = (y_1, y_2, \ldots y_L)$ related to the patient movements, we propose the following scoring function:

$$\bar{x}_k = \left(1 + e^{\frac{x_k}{\mu+3\delta} - \alpha_1}\right)^{-1}; \quad (4)$$

$$\bar{y}_k = \left(1 + e^{\frac{x_k}{(\mu+3\delta)} - \alpha_1} + \frac{y_k - x_k}{\alpha_2(\mu+3\delta)}\right)^{-1}, \quad (5)$$

where $k \in \underline{L}$, $\mu = \frac{1}{L}\sum_{k=1}^{L}|x_k|$, $\delta = \sqrt{\frac{1}{L}\sum_{k=1}^{L}(|x_k|-\mu)^2}$, and $\alpha_1$, $\alpha_2$ are data-specific parameters. The proposed scoring function is monotonically decreasing, and is designed to preserve the distribution of the values of the performance metric. The values for the reference movements $x_k$ are scaled by $\mu + 3\delta$ in (4) to ensure that the resulting scores $\bar{x}_k$ have values close to 1 for inputs $x_k$ in the range $(\mu - 3\delta, \mu + 3\delta)$. Similarly, for the patient movements $y_k$ the scoring function in (5) is designed to preserve their distribution in mapping the performance metric values into movement quality scores.

*E. Deep Learning Architecture for Rehabilitation Assessment*

We propose a novel deep learning model for spatio-temporal modeling of skeletal data, for application in rehabilitation assessment. A graphical representation of the NN architecture is provided in Fig. 3. The NN model is designed to exploit the spatial characteristics of human movements by dedicating sub-networks for processing joint displacements of individual body parts. In addition, the input data is arranged into temporal pyramids for processing multiple scaled version of the movement repetitions. The initial hierarchical layers in the model employ strided one-dimensional convolutional filters for learning spatial dependencies in human movements, and are followed by a series of LSTM recurrent layers for modeling temporal correlations in learned representations.

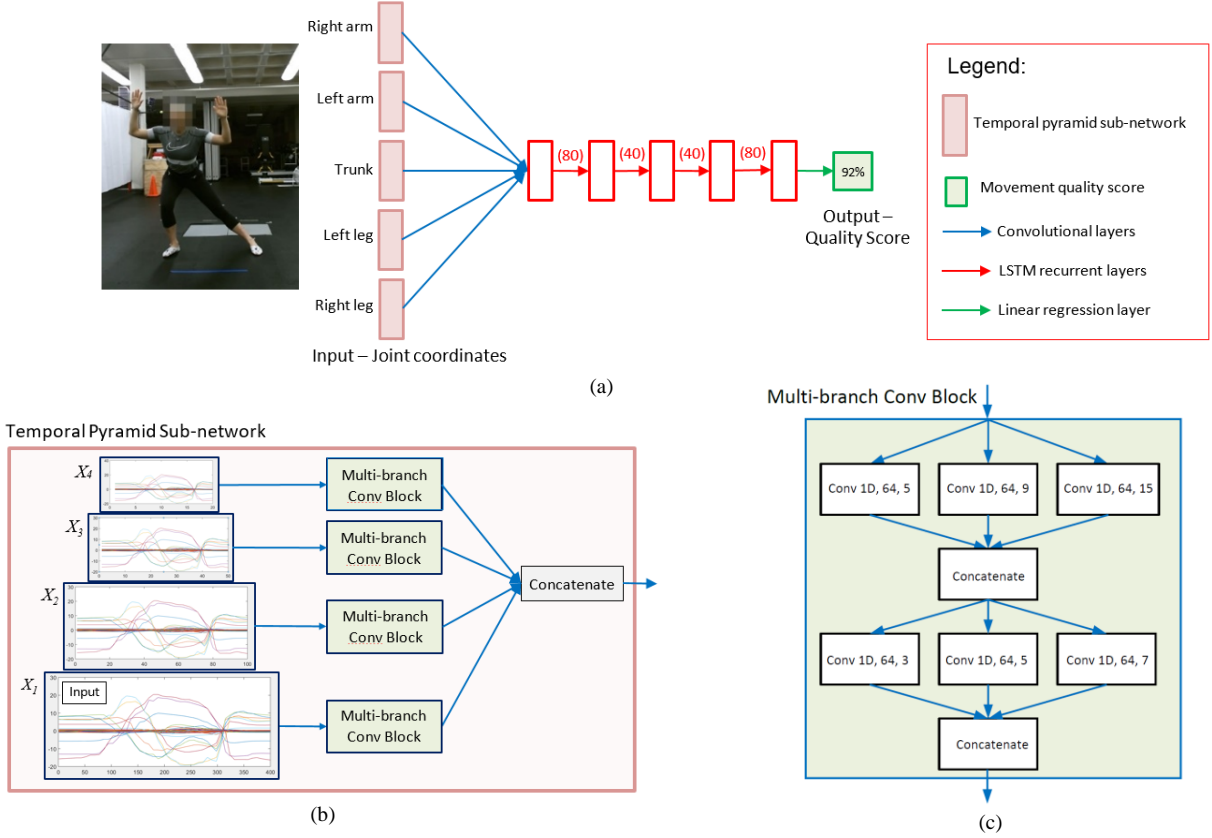

Fig 3. (a) The proposed spatio-temporal model architecture. (b) Temporal pyramid sub-network. (c) Multi-branch convolutional block.

The architecture of the NN draws inspiration from the hierarchical model proposed by Du et al. [10] that employs five recurrent sub-networks taking as inputs joint displacements of the left arm, right arm, left leg, right leg, and torso, respectively. The outputs from the five sub-networks are merged into a single representation. Such hierarchical organization of the network layers allows low-level spatial information from joint movements to be exploited for obtaining a high-level representation of the body parts' movements in accomplishing required actions. Differently form the model proposed by Du et al. [10] that consists of bidirectional layers with LSTM recurrent units, our proposed network uses convolutional units in the hierarchical layers and recurrent units in the succeeding layers. The presented ablation study and performance comparison in Section VI corroborate the advantage of the introduced modifications in our spatio-temporal model.

Similarly, the introduction of temporal pyramids for processing rehabilitation data was motivated by the concept of image pyramids in computer vision. Temporal pyramids have been used for processing video data by dynamically subsampling input videos at varying frame rate [43], temporal pooling of multi-scale data representations from extracted feature map [44], or by applying sliding windows with changeable scales to the sequences of images [45]. In these works, the use of multi-scale video pyramids has been conducive to improve the detection and localization of human actions in videos.

In the proposed network, the temporal pyramids are composed of full-scaled input sequences, and three sub-sampled versions with a temporal length equal to one half, one quarter, and one eight of the sequence (see Fig. 3(b)). The resulting feature vectors are then concatenated and passed to the next layers. Such data processing enables recognizing movement patterns at different levels of abstraction, and led to improved performance of the deep model for movement assessment.

Inputs to the network are 117-dimensional sequences of the full-body joint angles corresponding to single repetitions of an exercise. The convolutional blocks consist of two convolutional layers followed by dropout layers with a rate of 0.25. For these layers we adopted the multi-branch design approach shown in Fig. 3(c), popularized in the inception convolutional network architectures [46]. Each layer contains three branches of 1D convolutional filters with different length, which outputs are concatenated and passed to the next layer. The use of multiple branches allows the model the select the most suitable filter length based on the input data. The recurrent portion of the model consists of four layers with 80, 40, 40, and 80 LSTM units, respectively. The last layer has linear activations, and outputs a numerical movement quality score for an input repetition. Mean-squared-error was used as a cost function for training the model parameters, with the Adam optimizer. A batch size of 5 repetitions was applied, with early stopping regularization.

One can note that the proposed model is not particularly deep, as it comprises of a relatively low number of hidden layers; however, considering that the used dataset is also of relatively small size, larger and deeper networks would overfit and produce suboptimal results.

## IV. EXPERIMENTAL RESULTS

### A. Dataset

For validation of the presented framework, we created the UI-PRMD dataset [11]. The dataset consists of skeletal data collected from 10 healthy subjects. Each subject completed 10 repetitions of 10 rehabilitation exercises, listed in Table I. The data were acquired with a Vicon optical tracking system, and consist of 117-dimensional sequences of angular joint displacements. The subjects performed the exercises both in a correct manner, hereafter referred to as correct movements, and in an incorrect manner, i.e., simulating performance by patients with musculoskeletal constraints, hereafter referred to as incorrect movements. The research study related to the data collection was approved by the Institutional Review Boards at the University of Idaho under the identification code IRB 16-124. A written informed consent for participation in a research study was approved by the board, and was obtained from all participants in the study. A detailed description of the UI-PRMD dataset is provided in [11].

TABLE I
EXERCISES IN THE UI-PRMD DATASET

| Order | Exercise |
|---|---|
| E1 | Deep squat |
| E2 | Hurdle step |
| E3 | Inline lunge |
| E4 | Side lunge |
| E5 | Sit to stand |
| E6 | Standing active straight leg raise |
| E7 | Standing shoulder abduction |
| E8 | Standing shoulder extension |
| E9 | Standing shoulder internal–external rotation |
| E10 | Standing shoulder scaption |

### B. Performance Quantification

In this section, the adopted performance metric based on the log-likelihood of GMM is evaluated on the UI-PRMD dataset. For comparison, three common performance metrics for assessment of rehabilitation exercises based on Euclidean, Mahalanobis, and DTW distance are also evaluated.

*Data scaling:* To compare the performance metrics on the same basis, their values are first linearly scaled to the same range. In this study the range [1, 20] was selected based on an empirical understanding of the data. For the obtained values of the performance metrics related to repetitions of the correct movements denoted $\mathbf{x} = (x_1, x_2, \ldots, x_L)$, and for the metrics of the incorrect movements $\mathbf{y} = (y_1, y_2, \ldots, y_L)$, the following scaling functions were used

$$x_i' = \frac{19(x_i - m)}{M - m} + 1 \; ; \; y_i' = \frac{19(y_i - m)}{M - m} + 1 \text{ for } i \in \underline{L}, \quad (6)$$

where $x_i', y_i' \in [1, 20]$ denote the scaled values of the performance metrics, $M = \max_{i,j \in \underline{L}}\{x_i, y_j\}$, and $m = \min_{i,j \in \underline{L}}\{x_i, y_j\}$.

The scaled values of the Euclidean distance for exercises E1 and E2 are shown in Fig. 4. Green circle markers are used for the repetitions of the correct movements, whereas the red squares symbolize the repetitions of the incorrect movements. Note that inconsistent data (associated with measurement errors or subjects performing the exercise with their left-arm/leg in a set of mostly right arm/leg exercises) were manually removed from the original dataset, resulting in less than 100 repetitions per subject. E.g., there are 90 correct and incorrect movements for E1 in Fig. 4(a), and 55 correct and incorrect movements for E2 in Fig. 4(b).

*Separation degree:* For comparison of the scaled values of the performance metrics we propose the concept of separation degree. Specifically, for any positive real numbers $x, y$, their separation degree is defined as $S_D(x, y) = \frac{x-y}{x+y} \in [-1, 1]$. The separation degree between two positive sequences $\mathbf{x} = (x_1, x_2, \ldots, x_m)$ and $\mathbf{y} = (y_1, y_2, \ldots, y_n)$ is defined by

$$S_D(\mathbf{x}, \mathbf{y}) = \frac{1}{mn} \sum_{i=1}^{m} \sum_{j=1}^{n} S_D(x_i, y_j). \quad (7)$$

Values of the separation degree close to 1 or $-1$ indicate good separation between the two sequences. Conversely, for values of the separation degree close to 0, the sequences don't separate well and they are almost mixed together.

When applied to the values of the distance metrics, the separation degree indicates greater ability of the used metric to differentiate between correct and incorrect repetitions of an exercise. For instance, in Fig. 4(b) one can observe a clearer differentiation between the correct and incorrect movements, in comparison to Fig. 4(a). This results in a larger value of the separation degree for the repetitions of exercise E2, which were calculated at 0.384 for E1, and 0.497 for E2, respectively.

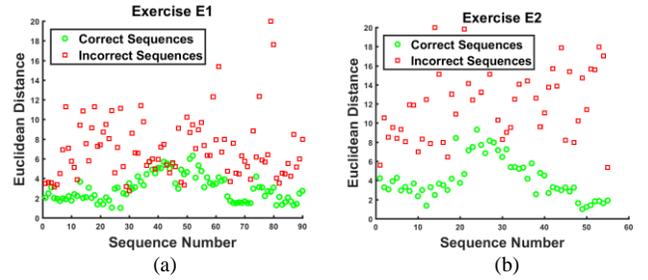

Fig. 4. Scaled values of the Euclidean distance for the between-subject case for: (a) First exercise E1 ( $S_D = 0.384$); (b) second exercise E2 ( $S_D = 0.497$).

The values for the separation degrees for the four studied performance metrics are presented in Table II. Each cell in the table corresponds to the average separate degree values $S_D$ for the 10 exercises in the dataset. The shown values are the means and in parentheses are the standard deviations. For the comparison, scaled values of the metrics according to (6) are used. Values for both between-subject and within-subject cases are presented. Table II also compares the values of the metrics for the cases of raw 117-dimensional data, and low-dimensional data obtained with the methods of maximum-variance, PCA, and GMM log-likelihood. The largest values for the separation degree are indicated in each row with a bold font.

Conclusively, the GMM log-likelihood metric applied on a low-dimensional data with the autoencoder NN resulted in the largest separation between the correct and incorrect movements for both between- and within-subject cases. The within-subject case provides improved separation because the repetitions performed by the same subject are characterized with a lower level of variability. The value of the GMM log-likelihood is not provided for the 117-dimensional data because GMM is commonly applied on low-dimensional data. Furthermore, the performance of the Euclidean and DTW distances in Table II is comparable, and better than the Mahalanobis distance. Also, one can notice that the autoencoder NN lost less information in compressing the high-dimensional data sequences in comparison to maximum variance and PCA, because the separation degree values for all metrics using autoencoders are very close to the corresponding metric values of the 117-dimensions data without dimensionality reduction. In implementing GMM on the dataset, the number of Gaussian components $C$ was set to 6.

TABLE II
SEPARATION DEGREE FOR THE PERFORMANCE METRICS: MEAN (ST. DEV.)

| Metric | Euclidean distance | Mahalanobis distance | DTW distance | Log-likelihood GMM |
|---|---|---|---|---|
| Between-subject | | | | |
| $D^a$=117 | 0.445 (0.087) | 0.195 (0.152) | **0.487** (0.063) | -- |
| D=3 (MV) | 0.309 (0.101) | 0.063 (0.130) | 0.310 (0.100) | **0.344** (0.049) |
| D=3 (PCA) | 0.296 (0.103) | 0.108 (0.169) | 0.265 (0.093) | **0.360** (0.060) |
| D=4 (AE) | 0.423 (0.092) | 0.229 (0.102) | 0.427 (0.094) | **0.515** (0.106) |
| Within-subject | | | | |
| D=117 | 0.568 (0.058) | 0.441 (0.118) | **0.570** (0.059) | -- |
| D=3 (MV) | **0.472** (0.048) | 0.325 (0.118) | 0.455 (0.053) | 0.471 (0.098) |
| D=3 (PCA) | 0.508 (0.032) | 0.322 (0.169) | 0.501 (0.031) | **0.518** (0.057) |
| D=4 (AE) | 0.582 (0.057) | 0.474 (0.133) | 0.574 (0.060) | **0.603** (0.073) |

D: data dimensions; MV: maximum variance; PCA: principal component analysis; AE: autoencoder neural networks

### C. Neural Networks Performance

For training the deep neural networks, the movement quality scores based on the GMM log-likelihood calculated with autoencoder-reduced data are employed. Only the case of between-subject is considered, since for the within-subject case the number of repetitions per subject is too small to train NNs.

*Scoring function:* The scoring function presented in (4)-(5) is used to calculate the movement quality scores. The values of the parameters are empirically selected as $\alpha_1 = 3.2$ and $\alpha_2 = 10$. For example, Fig. 5 depicts the values of the log-likelihood and the corresponding performance scores for exercise E1 (i.e., deep squat). The scores for the correct movements shown in Fig. 5(b) have values close to 1, whereas most of the scores for the incorrect movements are in the range between 0.7 and 0.9.

*NN evaluation*: The model was implemented on a Dell Precision 5810 workstation with Intel Xeon CPU, 32 GB RAM, 2 TB hard disk, and an NVIDIA Titan Xp GPU card. Inputs to the NNs are pairs of repetition data containing raw 117-dimensional angular joint measurements and quality scores. The networks are trained in a supervised regression manner, where the output is a predicted value of the movement quality for an input repetition. For each of the 10 exercises in the UI-PRMD dataset, a separate NN is trained and used for quality assessment. Each network model is run five times, and we report the average absolute deviation between the ground truth quality scores and the network prediction.

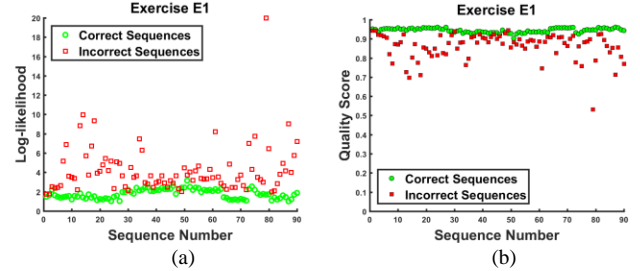

Fig 5. (a) GMM log-likelihood values for exercise E1; (b) Corresponding quality scores.

To evaluate the respective contributions of the individual components in the design of our spatio-temporal model we conducted an ablation study. The results for the 10 exercises in the dataset are displayed in Table III. Lower values of the absolute deviation indicate low errors by the NN model in predicting the quality score for input data. The upper row in the table presents the aggregated mean deviation for all exercises E1 to E10. The results of the ablation study support our intuitive assumptions that the introduced components in the proposed model related to the multi-branch layers, temporal pyramids, hierarchical structure, and combination of convolutional and recurrent units all contribute to improved assessment of rehabilitation exercises.

TABLE III
ABLATION STUDY: AVERAGE ABSOLUTE DEVIATION PER EXERCISE

| Exercise | Our Approach | Without Branching Layers | Without Temporal Pyramids | Without Hierarch. Layers | Without Recurrent Layers |
|---|---|---|---|---|---|
| E1-E10 | **0.02527** | 0.02537 | 0.02594 | 0.02953 | 0.04729 |
| E1 | **0.01077** | 0.01213 | 0.01162 | 0.01222 | 0.03631 |
| E2 | 0.02824 | **0.02415** | 0.02785 | 0.03522 | 0.04322 |
| E3 | **0.03980** | 0.04232 | 0.04286 | 0.05350 | 0.07876 |
| E4 | 0.01185 | 0.01495 | 0.01226 | **0.01048** | 0.03654 |
| E5 | 0.01870 | 0.01758 | **0.01569** | 0.01719 | 0.03716 |
| E6 | **0.01779** | 0.02110 | 0.01930 | 0.01858 | 0.04104 |
| E7 | **0.03819** | 0.03907 | 0.04241 | 0.04016 | 0.05699 |
| E8 | **0.02305** | 0.02369 | 0.02418 | 0.02658 | 0.04589 |
| E9 | **0.02271** | 0.02284 | 0.02296 | 0.02738 | 0.04130 |
| E10 | 0.04162 | **0.03584** | 0.04027 | 0.05395 | 0.05565 |

We further compared the performance of the proposed NN to state-of-the-art deep learning models for movement classification. We are not aware of any other deep NN models for movement assessment. On the other hand, there is a large body of research on using deep learning models for classification/recognition/detection of human movements (in a general context, rather than for biomedical purposes). Therefore, we adapted several recent NN classifiers that have achieved top performance, and we re-purposed the models for regressing movement quality scores. The selected models are: Co-occurrence [47], PA-LSTM [48], Two-stream CNN [49],

Hierarchical LSTM [10], as well as two basic Deep CNN and Deep LSTM architectures.

For these networks, we replaced the last softmax layer with a fully-connected layer with linear activations. Furthermore, we omitted all batch normalization layers (if any were present) in the original models, as they significantly degraded the capacity for movement assessment. Other than that, we closely followed the proposed implementation as described by the authors in the respective papers. Hierarchical LSTM is the network proposed by Du et al. [10] that served as a motivation for our proposed deep learning model. We selected the architectures and hyperparameters of the basic Deep CNN and Deep LSTM models through an extensive grid-search; the resulting CNN network has three convolutional layers (60, 30, and 10 units) followed by two fully-connected layers (200 and 100 units), whereas the Deep LSTM network contains one LSTM layer (20 units), one fully-connected layer (30 units), and another LSTM layer (10 units). The values of the average absolute deviations are presented in Table IV. With regards to the ability for movement quality assessment of all 10 exercises in the dataset, our proposed model outperformed the other deep learning classification models, although some of the models provided better performance on several of the exercises in the dataset (shown with a bold font in the table). The computational times for training the models averaged over all exercises are shown in the last row in Table IV. The proposed spatio-temporal model is computationally less expensive than almost all compared models. The prediction of movement quality scores for input repetitions by the trained model is very fast, and it took about 10 milliseconds per repetition on average.

TABLE IV
PERFORMANCE COMPARISON: AVERAGE ABSOLUTE DEVIATION PER EXERCISE

| Exercise | Our approach | Co-occurrence [47] | Deep CNN | PA-LSTM [48] | Two-stream CNN [49] | Hierar. LSTM [10] | Deep LSTM |
|---|---|---|---|---|---|---|---|
| E1-E10 | **0.02527** | 0.02703 | 0.02615 | 0.04534 | 0.11044 | 0.08819 | 0.04059 |
| E1 | 0.01077 | **0.01052** | 0.01357 | 0.01839 | 0.28798 | 0.03010 | 0.01670 |
| E2 | **0.02824** | 0.02905 | 0.02953 | 0.04413 | 0.22349 | 0.07742 | 0.04934 |
| E3 | **0.03980** | 0.05577 | 0.04141 | 0.08094 | 0.20493 | 0.13766 | 0.09382 |
| E4 | **0.01185** | 0.01347 | 0.01640 | 0.02347 | 0.36033 | 0.03580 | 0.01609 |
| E5 | 0.01870 | 0.01687 | **0.01300** | 0.03156 | 0.12332 | 0.06367 | 0.02536 |
| E6 | **0.01779** | 0.01886 | 0.02349 | 0.03426 | 0.21119 | 0.04676 | 0.02166 |
| E7 | 0.03819 | **0.02733** | 0.03346 | 0.04954 | 0.05016 | 0.19280 | 0.04090 |
| E8 | **0.02305** | 0.02464 | 0.02905 | 0.05070 | 0.04337 | 0.07260 | 0.04590 |
| E9 | **0.02271** | 0.02720 | 0.02495 | 0.04313 | 0.14411 | 0.06508 | 0.04419 |
| E10 | 0.04162 | 0.04657 | **0.03667** | 0.07727 | 0.11044 | 0.16009 | 0.05198 |
| Training time (in seconds) | | | | | | | |
| 177 | 325 | **52** | 598 | 4,668 | 295 | 410 | |

The results of the proposed deep NN for assessment of exercise E1 are depicted in Fig. 6. The set of 90 correct and 90 incorrect repetitions was randomly split using a ratio of 0.7/0.3 into a training set of 124 and a validation set of 56 repetitions. The ground truth scores and predicted scores for the training and validation sets are shown in Figs. 6(a) and (b), respectively. In the two sub-figures the first half of the scores are for the correct sequences and have values close to one, and the second half of the scores pertain to the incorrect sequences and have lower quality scores. Conclusively, the network predictions closely follow the values of the input quality scores for all data instances. We also validate the proposed approach using leave-one-out cross-validation (i.e., testing on one subject a model trained on all other subjects). The performance was comparable to the presented results using random test data, with the predicted quality scores closely following the ground truth values.

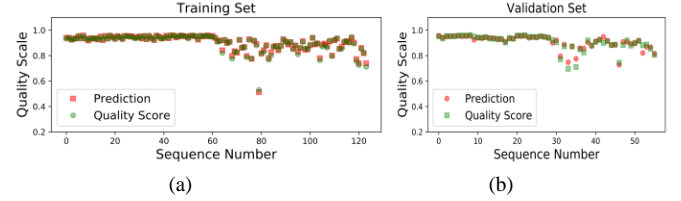

Fig. 6. (a) Predictions on the training set for exercise E1; (b) Predictions on the validation set for exercise E1.

The proposed model was next evaluated on the KIMORE dataset [50], which contains data for five rehabilitation exercises performed by 44 healthy subjects and 34 patients, and collected with a Kinect v2 sensor. We implemented our proposed deep learning model on the deep squat exercise. We employed full-body joint orientations data for 33 healthy subjects and 18 patients, and extracted 4 repetitions for each subject, resulting in 204 repetitions in total. The KIMORE dataset provides clinical scores for each subject's performance in the [0, 50] range. To train the model, we scaled the values in the [0, 1] range, and randomly selected 142 repetitions for training, and 62 for validation. The results are displayed in Figure 7, where the predicted movement quality scores by the deep learning model closely follow the ground truth scores provided by the clinicians. The obtained mean absolute deviation was 0.03786, which is greater than the deviation for the deep squat exercise in the UI-PRMD dataset, probably due to the lower accuracy of Kinect v2 compared to Vicon, and also in the KIMORE dataset the same clinical score is assigned for all repetitions performed by the same subject.

## V. DISCUSSION

The article introduces a novel framework for the assessment of rehabilitation exercises via deep NNs. The framework includes performance metrics, scoring functions, and NN models. Common metrics for quantifying the level of consistency in captured rehabilitation movements are compared. The metrics include Euclidean, Mahalanobis, DTW distance, and GMM log-likelihood. The concept of separation degree is proposed for metric comparison. GMM log-likelihood outperformed the model-less metrics on the UI-PRMD dataset. Such results confirm our hypothesis that efficient movement assessment is strongly predicated on the provision of models of human movements. Probabilistic approaches, such as the used GMM approach, have improved ability to handle the inherent variability and measurement uncertainty in human movement data, in comparison to the model-less approaches.

We compared the performance of PCA and maximum variance approaches for dimensionality reduction of human movements to autoencoder NNs. Expectedly, the provision of nonlinear functions for neuron activations in autoencoders provides richer representational capacity of the data into a lower dimensional space, in comparison to the linear technique of PCA and the simple concept of maximum variance.

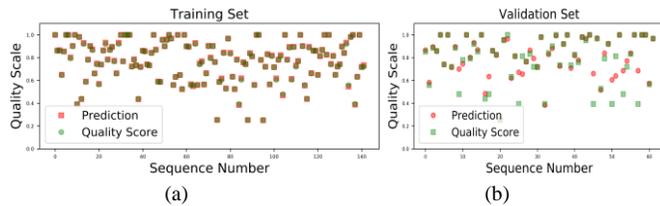

Fig. 7. (a) Predictions on the training set for deep squat exercise; (b) Predictions on the validation set for deep squat exercise.

We propose a deep learning architecture for hierarchical spatio-temporal modeling of rehabilitation exercises at multiple levels of abstraction. NNs are trained for each exercise via supervised regression, where for inputs comprising exercise repetitions the inferred outputs are quality scores. The network structure combines hierarchical merging of extracted feature vectors from different body parts, pyramidal processing of the movement sequences subsamples at multiple temporal scales, and multi-branch blocks for learning the structure of the used computational units. Although recurrent units are most commonly used for processing sequential time-series data as the considered rehabilitation movements, our proposed model employs convolutional filters in the initial layers and LSTM recurrent units in the later layers of the network. The reasons for such design stem from the following: (1) the employed dataset is fairly small, consisting of less than 200 repetitions per exercises, hence recurrent NNs can overfit the data due to the larger number of used parameters, and (2) a growing body of work report of improved performance by CNNs on time-series and movement data [51]. The proposed deep learning model outperformed recent state-of-the-art deep NNs designed for movement classification.

Our presented research has several limitations. The validation is primarily performed on rehabilitation exercises performed by healthy subjects, where the measurements are acquired with an expensive optical motion capturing system. Additionally, the largest segment of the validation is based on movement data without a ground truth assessment of the movement quality by clinicians. The evaluation of the deep squat exercise in the KIMORE dataset provides a partial validation on patient data collected with a low-cost sensor.

In future work, we will attempt to address the above-listed shortcomings of this study, i.e., we will focus on a thorough validation of the framework on rehabilitation exercises performed by patients and labeled by a group of clinicians who will assign quality scores. We will validate the proposed approach by acquiring muscle activity measurements. Also, we have plans to implement the framework for assessment of patient performance in home-based rehabilitation using a Kinect sensor.

## VI. CONCLUSION

The article proposes a deep learning-based framework for assessment of rehabilitation exercises. The framework consists of algorithms for dimensionality reduction, performance metrics, scoring functions, and deep learning models. The framework is evaluated on a dataset of 10 rehabilitation exercises. The experimental results indicate that the quality scores generated by the proposed framework closely follow the ground truth quality scores for the movements.

This work demonstrates the potential of deep learning models for assessment of rehabilitation exercises. Such models can consistently outperform the approaches that employ distance functions for movement assessment where the data processing is performed on low-level measurements of joint coordinates at the individual time-steps, and the probabilistic approaches where the data modeling is typically performed at a single level of abstraction. The advantages of deep NNs for this task originate from the capacity for hierarchical modeling of human movements at multiple spatial and temporal levels of abstraction. This type of models provide improved abilities to "understand" the levels of hierarchy and the complex spatiotemporal correlations in human movement data.